\pdfpageattr {/Group << /S /Transparency /I true /CS /DeviceRGB>>}
\documentclass{article}

\usepackage[final]{nips_2016}

\usepackage[utf8]{inputenc} 
\usepackage[T1]{fontenc}    
\usepackage{hyperref}       
\usepackage{url}            
\usepackage{booktabs}       
\usepackage{amsfonts}       
\usepackage{nicefrac}        
\usepackage{microtype}      

\author{
Stephan Zheng\\
Caltech\\
\tt{stzheng@caltech.edu}
\And
Yisong Yue\\
Caltech\\
\tt{yyue@caltech.edu}
\And
Patrick Lucey\\
STATS\\
\tt{plucey@stats.com}
}

\usepackage{wrapfig}
\usepackage[font={small,it}]{caption}

\usepackage{natbib}

\usepackage{graphicx}
\graphicspath{{./images/}}
\DeclareGraphicsExtensions{.pdf,.jpeg,.png}

\usepackage{subcaption}
\newcommand{\History}[1]{h_{#1}}
\newcommand{\GoalState}{g}
\newcommand{\GameState}{s}
\newcommand{\FullAction}{a}
\newcommand{\MicroAction}{a}

\newcommand{\AuxAction}{u}

\newcommand{\MacroPol}{\pi_{\trm{macro}}}
\newcommand{\MicroPol}{\pi_{\trm{micro}}}
\newcommand{\AuxPol}{\pi_{\trm{raw}}}

\newcommand{\MacroP}{P^{\trm{macro}}}
\newcommand{\MicroP}{P^{\trm{micro}}}
\newcommand{\AuxP}{P^{\trm{raw}}}

\newcommand{\transferF}{\phi}
\newcommand{\TransferF}{\Phi}
\newcommand{\synthesisF}{\psi}
\newcommand{\SynthesisF}{\Psi}

\newcommand{\attwidth}{0.23\textwidth}
\newcommand{\attheight}{0.22\textwidth}
\newcommand{\HPN}{HPN }
\newcommand{\micpl}{\pi_\trm{micro}}
\newcommand{\macpl}{\pi_\trm{macro}}

\newcommand{\cnn}{\tsc{cnn} }
\newcommand{\grucnn}{\tsc{gru-cnn} }

\newcommand{\grucnnhiercc}{\tsc{h-gru-cnn-cc} }
\newcommand{\grucnnhieraux}{\tsc{h-gru-cnn-aux} }
\newcommand{\grucnnhierstack}{\tsc{h-gru-cnn-stack} }
\newcommand{\grucnnhierfull}{\tsc{h-gru-cnn-att}}

\newcommand{\INST}[1]{}

\newcommand{\KL}[1]{\textbf{KL}}

\newcommand{\tempnewpage}{}

\usepackage{subcaption}

\usepackage[normalem]{ulem}
\usepackage[ruled]{algorithm}
\usepackage{algpseudocode}
\usepackage{amsmath, amsthm, amssymb}
\usepackage{array} 
\usepackage{caption}
\usepackage{float}
\usepackage{framed}
\usepackage{gensymb} 
\usepackage{mathtools}
\usepackage{multirow}
\usepackage{subcaption}
\usepackage{wrapfig}
\usepackage{blindtext}

\usepackage{color}
\definecolor{blue}{rgb}{0,0,.7}
\definecolor{red}{rgb}{.7,0,0}
\definecolor{orange}{rgb}{1,.6,0}
\definecolor{purple}{rgb}{.4,0,.5}
\definecolor{brown}{rgb}{.4,.2,.1}
\definecolor{green}{rgb}{0,.5,0}

\newcommand{\refn}[1]{(\ref{#1})}

\newcommand{\brck}[1]{\ensuremath{\left(#1\right)}}
\newcommand{\brcksq}[1]{\ensuremath{\left[#1\right]}}
\newcommand{\brckcur}[1]{\ensuremath{\left\{#1\right\}}}

\newcommand{\be}{\begin{equation}}
\newcommand{\ee}{\end{equation}}
\newcommand{\bali}{\begin{eqnarray*}}
\newcommand{\eali}{\end{eqnarray*}}
\newcommand{\eq}[1]{\begin{align}#1\end{align}}

\newcommand{\iitem}[1]{\begin{itemize}#1\end{itemize}}
\newcommand{\enumm}[1]{\begin{enumerate}#1\end{enumerate}}

\newcommand{\calA}{\mathcal{A}}

\newcommand{\calG}{\mathcal{G}}

\newcommand{\calS}{\mathcal{S}}

\newcommand{\trm}[1]{\textnormal{#1}}
\newcommand{\tsc}[1]{\textsc{#1}}
\newcommand{\tb}[1]{\textbf{#1}}
\newcommand{\ti}[1]{\textit{#1}}

\title{Generating Long-term Trajectories Using Deep Hierarchical Networks}

\begin{document}

\maketitle

\begin{abstract}
We study the problem of modeling spatiotemporal trajectories over long time horizons using expert demonstrations. For instance, in sports, agents often choose action sequences with long-term goals in mind, such as achieving a certain strategic position.
Conventional policy learning approaches, such as those based on Markov decision processes, generally fail at learning cohesive long-term behavior in such high-dimensional state spaces, and are only effective when myopic modeling lead to the desired behavior. The key difficulty is that conventional approaches are ``shallow'' models that only learn a single state-action policy.
We instead propose a hierarchical policy class that automatically reasons about both long-term and short-term goals, which we instantiate as a hierarchical neural network.
We showcase our approach in a case study on learning to imitate demonstrated basketball trajectories, and show that it generates significantly more realistic trajectories compared to non-hierarchical baselines as judged by professional sports analysts.
\end{abstract}
\section{Introduction}

\begin{wrapfigure}{r}{0.25\textwidth}
\vspace{-0.18in}
\centering
\includegraphics[trim={30px 25px 5px 5px},clip, width=0.25\textwidth]{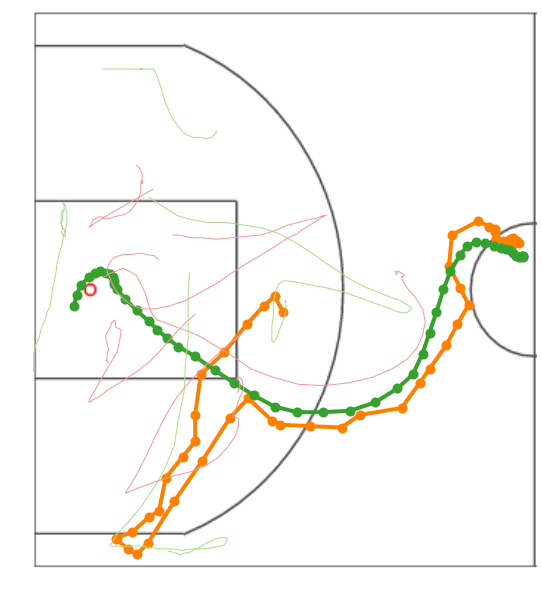}
\vspace{-0.2in}
\caption{The player (green) has two macro-goals: 1) pass the ball (orange) and 2) move to the basket.}\vspace{-15pt}
\label{fig:intro}
\end{wrapfigure}
Modeling long-term behavior is a key challenge in many learning problems that require complex decision-making.
Consider a sports player determining a movement trajectory to achieve a certain strategic position. The space of all such trajectories is prohibitively large, and precludes conventional approaches, such as those based on Markovian dynamics.

Many decision problems can be naturally modeled as requiring high-level, long-term \ti{macro-goals}, which span time horizons much longer than the timescale of low-level \ti{micro-actions} (cf. \cite{he_puma:_2010,ICLR16-hausknecht}).
A natural example for such macro-micro behavior occurs in spatiotemporal games, such as basketball where players execute complex trajectories. The micro-actions of each agent are to move around the court and if they have the ball, dribble, pass or shoot the ball. These micro-actions operate at the millisecond scale, whereas their macro-goals, such as "maneuver behind these 2 defenders towards the basket", span multiple seconds.
Figure \ref{fig:intro} depicts an example from a professional basketball game, where the player must make a sequence of movements (micro-actions) in order to reach a specific location on the basketball court (macro-goal).

Intuitively, agents need to trade-off between short-term and long-term behavior: often sequences of individually reasonable micro-actions do not form a cohesive trajectory towards a macro-goal. For instance, in Figure \ref{fig:intro} the player (green) takes a highly non-linear trajectory towards his macro-goal of positioning near the basket. As such, conventional approaches are not well suited for these settings, as they generally use a single (low-level) state-action planner, which is only successful when myopic or short-term decision-making leads to the desired behavior.

In this paper, we propose a novel class of \ti{hierarchical policy models}, which we instantiate using recurrent neural networks, that can simultaneously reason about both macro-goals and micro-actions. Our model utilizes an attention mechanism through which the macro-planner guides the micro-planner. Our model is further distinguished from previous work on hierarchical policies by dynamically predicting macro-goals instead of following fixed goals, which gives additional flexibility to our model class that can be fitted to data (rather than having the macro-goals be hand-crafted).

We showcase our approach in a case study on learning to imitate demonstrated behavior in professional basketball.
Our primary result is that our approach generates significantly more realistic player trajectories compared to non-hierarchical baselines, as judged by professional sports analysts. We also provide a comprehensive qualitative and quantitive analysis, e.g., showing that incorporating macro-goals can actually improve 1-step micro-action prediction accuracy. \tempnewpage
\section{Related Work}
\label{sec:related}

The reinforcement learning community has largely focused on non-hierarchical, or ``shallow'', policies such as those based on Markovian or linear dynamics (cf. \cite{ziebart2008maximum,mnih_human-level_2015,ICLR16-hausknecht}). By and large, such policy classes are only shown to be effective when the optimal action can be found via short-term planning. Previous research has instead focused on issues such as how to perform effective exploration, plan over parameterized action spaces, or deal with non-convexity issues from using deep neural networks within the policy model. In contrast, we focus on developing hierarchical policices that can effectively generate realistic long-term plans in complex settings such as basketball gameplay.

The use of hierarchical models to decompose macro-goals from micro-actions is relatively common in the planning community (cf. \cite{sutton_between_1999,he_puma:_2010,bai2015online}). For instance, the winning team in 2015 RoboCup Simulation Challenge \cite{bai2015online} used a manually constructed hierarchical planner to solve MDPs with a set of fixed sub-tasks. In contrast, we study how to \textit{learn} a hierarchical planner from a large amount of expert demonstrations that can adapt its policy in non-Markovian environments with dynamic macro-goals.

A somewhat related line of research aims to develop efficient planners for factored MDPs \cite{guestrin_efficient_2003}, e.g., by learning value functions over factorized state spaces for multi-agent systems. It may be possible that such a factorization approach is also applicable for learning our hierarchical structure. However, it is less clear how to combine such a factorization method with utilizing deep networks.

Our hierarchical model bears affinity to other attention models for deep networks that have mainly been applied to natural language processing, image recognition and combinations thereof \cite{xu_show_2015}. In contrast to previous work which focuses on attention models of the input, our attention model is applied to the \textit{output} by integrating control from both the macro-planner and the micro-planner.

Recent work on generative models for sequential data \cite{NIPS2015_5653}, such as handwriting generation, have combined latent variables with an RNN's hidden state to capture temporal variability in the input. In our work, we instead aim to learn semantically meaningful latent variables that are external to the RNN and reason about long-term behavior and goals.

Our model shares conceptual similarities to the Dual Process framework \cite{evans_dual-process_2013}, which decomposes cognitive processes into fast, unconscious behavior (System 1) and slow, conscious behavior (System 2). This separation reflects our policy decomposition into a macro and micro part. Other related work in neuroscience and cognitive science include hierarchical models of learning by imitation \cite{byrne1998learning}.\tempnewpage
\section{Long-Term Trajectory Planning}
\label{sec:theory}

We are interested in learning policies that can produce high quality trajectories, where quality is some global measure of the trajectory (e.g., realistic trajectories as in Figure \ref{fig:intro}). We first set notation:
\iitem{
\item Let $\calS, \calA$ denote the state, action space. Macro policies also use a goal space $\calG$.
\item At time $t$, an agent $i$ is in state $s^i_t\in\calS$ and takes action $\MicroAction^i_t\in\calA$. The state and action vectors are $\GameState_t = \brckcur{s^i_t}_{\trm{players } i}$, $\FullAction_t = \brckcur{\MicroAction^i_t}_{\trm{players } i}$.
The history of events is $\History{t} = \brckcur{(s_u,a_u)}_{0 \leq u < t}$.
\item Let $\pi(s_t,\History{t})$ denote a policy that maps state and history to a distribution over actions $P(a_t|s_t,\History{t})$. If $\pi$ is deterministic, the distribution is peaked around a specific action. We also abuse notation to sometimes refer to $\pi$ as deterministically taking the most probable action $\pi(s_t,\History{t}) = \trm{argmax}_{a\in\calA} P(a|s_t,\History{t})$ -- this usage should be clear from context.
}

Our main research question is how to design a policy class that can capture the salient properties of how expert agents execute trajectories. In particular, we present a general policy class that utilizes a goal space $\calG$ to guide its actions to create such trajectory histories. We show in Section \ref{sec:model} how to instantiate this policy class as a hierarchical network that uses an attention mechanism to combine macro-goals and micro-actions.
In our case study on modeling basketball behavior (Section \ref{sec:experiments}), we train such a policy to imitate expert demonstrations using a large dataset of tracked basketball games.

\subsection{Incorporating Macro-Goals}
\begin{wrapfigure}{r}{0.2\textwidth}
\vspace{-18pt}
 \begin{center}
  \includegraphics[trim={20px 15px 0px 10px}, clip, width=0.2\textwidth]{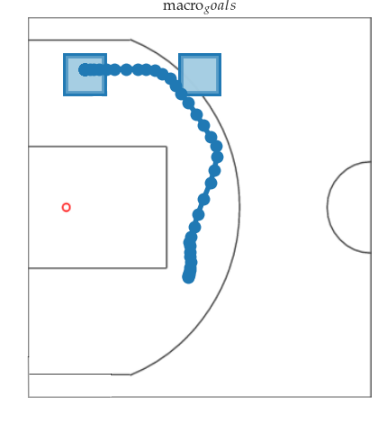}\hspace{-1pt}
 \end{center}
 \vspace{-9pt}
 \caption{Depicting two macro-goals (blue boxes) as an agent moves to the top left.}
 \label{fig:macrogoal-explanation}
\vspace{-20pt}
\end{wrapfigure}
Our main modeling assumption is that a policy \ti{simultaneously optimizes behavior hierarchically on multiple well-separated timescales}. We consider two distinct timescales (\ti{macro} and \ti{micro}-level), although our approach could in principle be generalized to even more timescales.
During an episode $[t_0, t_1]$, an agent $i$ executes a sequence of micro-actions $\brck{\MicroAction^i_t}_{t\geq 0}$ that leads to a macro-goal $\GoalState^i_t\in\calG$. We do not assume that the start and end times of an episode are fixed. For instance, macro-goals can change before they are reached.
We finally assume that macro-goals are relatively static on the timescale of the micro-actions, that is: $d\GoalState^i_t / dt \ll 1$.

Figure \ref{fig:macrogoal-explanation} depicts an example of an agent with two unique macro-goals over a 50-frame trajectory.
At every timestep $t$, the agent executes a micro-action $\MicroAction^i_t$, while the macro-goals $g^i_t$ change more slowly.

\begin{figure}[t]
\centering
\includegraphics{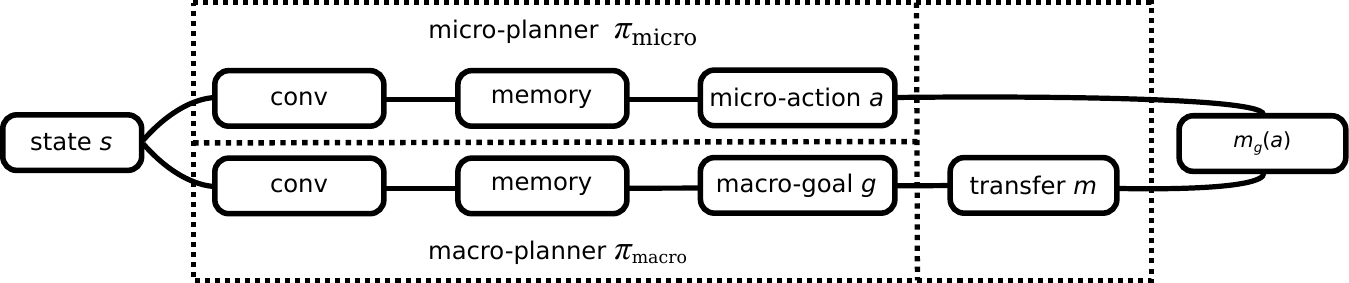}
\caption{The general structure of a 2-level hierarchical policy that consists of 1) a raw micro-policy $\AuxPol$ 2) a macro-policy $\MacroPol$ 3) a transfer function $m$ and 4) a synthesis function $\SynthesisF$. The raw micro-policy learns optimal short-term policies, while the macro-policy is optimized to achieve long-term rewards. The macro-policy guides the raw micro-policy $u = \AuxPol(s_t,h_t)$ in order for the hierarchical policy $\MicroPol = \synthesisF(u, \transferF(g))$ to achieve a long-term goal $g$. For a state $s_t$, the raw micro-policy outputs an action $\AuxPol(s_t,h_t)\in\calA$, while the macro-policy outputs a macro-goal $g_t=\MacroPol(s_t,h_t)$. The hierarchical policy then outputs an action $\MicroPol$ using the joint distribution over the raw actions and macro-goals, as defined by the transfer function $\transferF$ and synthesis functon $\synthesisF$.}
\label{fig:policynetwork_highlevel}
\vspace{-1pt}
\end{figure}

We model the interaction between a micro-action $\MicroAction^i_t$ and a macro-goal $\GoalState^i_t$ through a raw micro-action $\AuxAction^i_t\in\calA$ that is independent of the macro-goal. The micro-policy is then defined via a joint distribution over the raw micro-action and macro-goal:
\eq{\label{eq:micmacraw}
\MicroAction^i_t &= \MicroPol(\GameState^i_t, \History{t}^i) = \trm{argmax}_a \MicroP(a|\GameState^i_t, \History{t}^i) \\
\MicroP(\MicroAction^i_t | \GameState^i_t, \History{t}^i) &= \int d\AuxAction d\GoalState P(\MicroAction^i_t | \AuxAction, \GoalState) P(\AuxAction, \GoalState | \GameState^i_t, \History{t}^i).\label{eq:micmacraw2}
}
Here, we model the conditional distribution $P(a|u,g)$ as a non-linear factorization:
\eq{\label{eq:a_cond_ug}
  \MicroAction^i_t = \synthesisF(\AuxAction^i_t, \transferF(\GoalState^i_t)), \hspace{5pt} P(\MicroAction^i_t | \AuxAction^i_t, \GoalState^i_t) = \SynthesisF\brck{\AuxP(\AuxAction^i_t), \TransferF\brck{ \MacroP(\GoalState^i_t)} },
}
where $\transferF, \TransferF$ are non-linear transfer functions and $\synthesisF, \SynthesisF$ are synthesis functions - we will elaborate on the specific choices of $\TransferF, \SynthesisF$ in Section \ref{sec:model}. In this work, we do not condition \refn{eq:a_cond_ug} explicitly on $\GameState_t, \History{t}$, as we found this did not make a significant difference in our experiments. This hierarchical decomposition is visualized in Figure \ref{fig:policynetwork_highlevel} and can be generalized to multiple scales $l$ using multiple macro-goals $g^l$ and transfer functions $\transferF^l$.

 \tempnewpage
\section{Hierarchical Policy Network}
\label{sec:model}
Figure \ref{fig:policynetwork_highlevel} depicts a high-level overview of our hierarchical policy class for long-term spatiotemporal planning. In this work, both the raw micro-policy and macro-policy are instantiated as recurrent convolutional neural networks, and the raw action and macro-goals are combined through an attention mechanism, which we elaborate on below.

\tb{Attention mechanism for integrating macro-goals and micro-actions.} %

We model the joint distribution to implement an \ti{attention mechanism over the action space $\calA$}, that is, a non-linear weight function on $\calA$. For this, $\SynthesisF = \odot$ is a Hadamard product and the transfer function $\transferF$ maps a macro-goal $g$ into an attention variable $\transferF(g)$, whose distribution acts as an attention.
Suppressing indices $i, t$ for clarity, the conditional distribution \refn{eq:a_cond_ug} becomes
\eq{
P(\MicroAction | \AuxAction, \GoalState) = \delta_{a,u,\phi(g)} \AuxP(u|s,h) \odot \TransferF\brck{ \MacroP(g|s,h) } \hspace{5pt} (\trm{Hadamard}),
}
Intuitively, this structure captures the trade-off between the macro- and raw micro-policy.
On the one hand, the raw micro-policy $\AuxPol$ aims for short-term optimality,
On the other hand, the macro-policy $\MacroPol$ can attend via $\TransferF$ to sequences of actions that lead to a macro-goal and bias the agent towards good long-term behavior.
The difference between $u$ and $\transferF(g)$ thus reflects the trade-off that the hierarchical policy learns between actions that are good for either short-term or long-term goals.

\tb{Discretization and deep neural architecture.} In general, when using continuous (stochastic) latent variables $g$, learning the model \refn{eq:micmacraw} is intractable, and one would have to resort to approximation methods. Hence, to reduce complexity we discretize the state-action and latent spaces.
In spatiotemporal settings, such as for basketball, a state $s_t^i\in\calS$ can naturally be represented as a 1-hot occupancy vector of the basketball court. We then pose goal states $\GoalState^i_t$ as sub-regions of the court that $i$ wants to reach, defined at a coarser resolution than $\calS$.
With this choice, it is natural to instantiate the macro and micro-policies as convolutional recurrent neural networks, which can capture both predictive spatial patterns, non-Markovian temporal dynamics and implement the non-linear attention $\Phi$.

\tb{Multi-stage learning.} Given a training set $D$ of sequences of state-action tuples $(\GameState_t, \hat{a}_{t})$, where the $\hat{a}_{t}$ are 1-hot labels, the hierarchical policy network can be trained by solving an optimization problem
\eq{
\theta^* = \underset{\theta}{\trm{argmin}} \sum_{D}\sum_{t=1}^T L_t(\GameState_t, \History{t}, \hat{a}_{t}; \theta).\label{eq:argmin}
}
Given the hierarchical structure of our model class, we decompose the loss $L_t$ (dropping the index $t$):
\eq{
L(\GameState, \History{}, \hat{a}; \theta) &=
L_{\trm{macro}}\brck{\GameState, \History{}, \GoalState; \theta} +
L_{\trm{micro}}\brck{\GameState, \History{}, \hat{a}; \theta} + R(\theta),\label{eq:opt_highlevel}\\
L_{\trm{micro}}(\GameState, \History{}, \hat{a}; \theta) &= \sum_{k=1}^{A} \hat{a}_{k}%
\log\brcksq{ \AuxP(u_{k}|\GameState, \History{};\theta) \cdot \TransferF(\transferF(\GoalState)_{k}|\GameState, \History{};\theta)  }, \label{eq:loss-macromicro}
}
where $R_t(\theta)$ regularizes the model weights $\theta$ and $k$ indexes $A$ discrete action-values. Although we have ground truths $\hat{a}_t$ for the observable micro-actions, in general we do not have labels for the macro-goals $g_t$ that induce optimal long-term planning. As such, one would have to appeal to separate solution methods to compute the posterior $P(\GoalState_t|\GameState_t, \History{t})$ which minimizes $L_{t, \trm{macro}}\brck{\GameState_t, \History{t}; \GoalState_t, \theta}$.

To reduce this complexity and given the non-convexity of \refn{eq:loss-macromicro}, we instead follow a multi-stage learning approach with a set of \ti{weak labels} $\hat{\GoalState}_t, \hat{\transferF}_t$ for the macro-goals $g_t$ and attention masks $\transferF_t = \transferF(\GoalState_t)$. We assume access to such weak labels and only use them in the initial training phases. In this approach, we first train the raw micro-policy, macro-policy and attention individually, freezing the other parts of the network. The policies $\MicroPol, \MacroPol$ and attention $\transferF$ can be trained using standard cross-entropy minimization with the labels $\hat{a}_t, \hat{g}_t$ and $\hat{\transferF}_t$, respectively. In the final stage we fine-tune the entire network on objective \refn{eq:argmin}, using only $L_{t,\trm{micro}}$ and $R$. We found this approach beneficial to find a good initialization for the fine-tuning phase and to achieve high-quality long-term planning performance.%
\footnote{As $u_t$ and $\transferF(\GoalState_t)$ enter symmetrically into the objective \refn{eq:loss-macromicro}, it is possible, in principle, that the network converges to a symmetric phase where the predictions $u_t$ and $\transferF(\GoalState_t)$ become identical along the entire trajectory. However, our experiments suggest that our multi-stage learning approach separates timescales well between the micro- and macro policy and prevents the network from settling in the redundant symmetric phase.}
Another advantage of this approach is that we can train the model with gradient descent in all stages. \tempnewpage
\begin{figure}[t]
\centering
\includegraphics[width=\textwidth,height=0.22\textwidth]{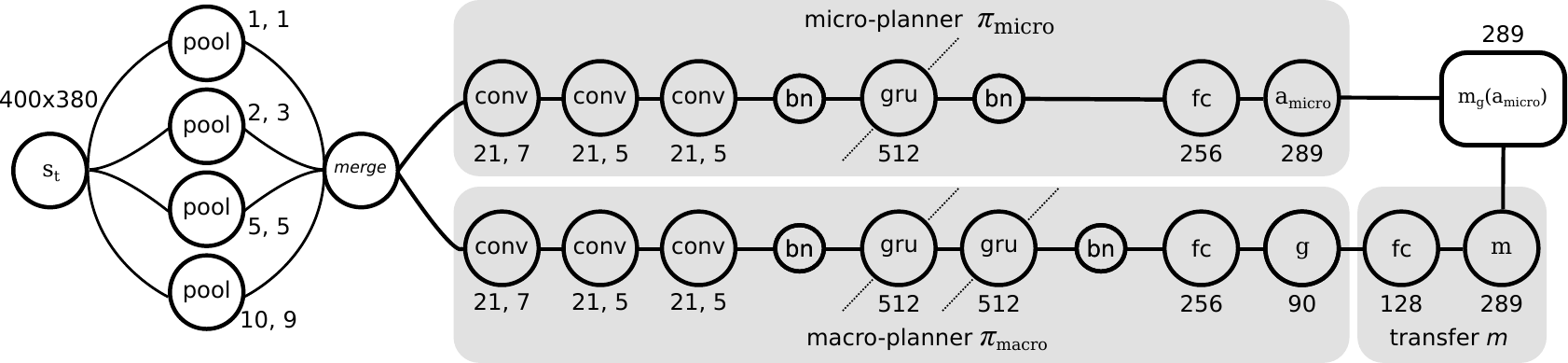}
\caption{Network architecture and hyperparameters of the hierarchical policy network, displayed for a single timestep $t$. Max-pooling layers (numbers indicate kernel size) with unit stride upsample the sparse tracking sequence data $\GameState_t$. Both the micro and macro planners $\micpl, \macpl$ use a convolutional (kernel size, stride) and GRU memory (number of cells) stack to predict $a_t$ and $g_t$. Batch-normalization (bn) \cite{ioffe_batch_2015} is applied to stabilize training. The output attention mechanism $m$ is implemented by 2 fully-connected layers (number of output units). Finally, the network predicts the final output $\MicroPol(s_t, h_t) = \AuxPol(s_t, h_t) \odot \transferF(g_t) $.
}
\label{fig:network_spec}
\end{figure}

\section{Case Study on Modeling Basketball Behavior}
We applied our approach to modeling basketball behavior data. In particular, we focus on imitating the players' movements, which is a challenging problem in the spatiotemporal planning setting.
\subsection{Experimental Setup}
\label{sec:experiments}
We validated the hierarchical policy network (HPN) by learning a basketball movement policy that predicts as the micro-action the \ti{instantaneous velocity} $v^i_t = \MicroPol(\GameState^i_t, \History{t}^i)$.

\tb{Training data.}
We trained the \HPN on a large dataset of tracking data from professional basketball games \cite{Yue2014}.
\footnote{A version of the dataset is available at \url{https://www.stats.com/data-science/}.}
The dataset consists of \ti{possessions} of variable length: each possession is a sequence of tracking coordinates $s^i_t = \brck{x^i_t, y^i_t}$ for each player $i$, recorded at 25 Hz, where one team has continuous possession of the ball.
Since possessions last between 50 and 300 frames, we sub-sampled every 4 frames and used a fixed input sequence length of 50 to make training feasible. Spatially, we discretized the left half court using $400\times380$ cells of size 0.25ft $\times$ 0.25ft.
For simplicity, we modeled every player identically using a single policy network. 
The resulting input data for each possession is grouped into 4 channels: the ball, the player's location, his teammates, and the opposing team.
After this pre-processing, we extracted 130,000 tracks for training and 13,000 as a holdout set.

\tb{Labels.}
We extracted micro-action labels $\hat{v}^i_t = s^i_{t+1}-s^i_t$ as 1-hot vectors in a grid of $17\times17$ unit cells.
Additionally, we constructed a set of weak macro-labels $\hat{\GoalState}_t, \hat{\transferF}_t$ by heuristically segmenting each track using its stationary points. The labels $\hat{\GoalState}_t$ were defined as the next stationary point. For $\hat{\transferF}_t$, we used 1-hot velocity vectors $v^i_{t,\trm{straight}}$ along the straight path from the player's location $s^i_t$ to the macro-goal $\GoalState^i_t$.
We refer to the supplementary material for additional details.

\tb{Model hyperparameters.} To generate smooth rollouts while sub-sampling every 4 frames, we simultaneously predicted the next $4$ micro-actions $a_{t}, \ldots, a_{t+3}$.
A more general approach would model the dependency between look-ahead predictions as well, e.g. $P(\pi_{t+\Delta+1}|\pi_{t+\Delta})$. However, we found that this variation did not outperform baseline models.
We selected a network architecture to balance performance and feasible training-time. The macro and micro-planner use GRU memory cells \cite{ChungGCB15} and a memory-less 2-layer fully-connected network as the transfer function $m$, as depicted in Figure \ref{fig:network_spec}. We refer to the supplementary material for more details. 

\tb{Baselines.}
We compared the \HPN against two natural baselines.
\enumm{
\item A policy with only a raw micro-policy and memory (\tsc{gru-cnn}) and without memory (\tsc{cnn}).
\item A hierarchical policy network \grucnnhiercc without an attention mechanism, which instead learns the final output from a concatenation of the raw micro- and macro-policy.
}

\tb{Rollout evaluation.} To evaluate the quality of our model, we generated rollouts $\brck{s_t; \History{0,r_0}}$ with \ti{burn-in period} $r_0$, These are generated by 1) feeding a ground truth sequence of states $\History{0,r_0}=\brck{\GameState_0, \ldots, \GameState_{r_0}}$ to the policy network and 2) for $t>r_0$, iteratively predicting the next action $a_t$ and updating the game-state $\GameState_t \rightarrow \GameState_{t+1}$, using ground truth locations for the other agents.

 \tempnewpage
\begin{figure}[!t]
\centering

\begin{subfigure}[b]{.19\textwidth}
 \centering
\includegraphics[trim={20px 15px 0px 10px}, clip, width=\textwidth, height=1.1\textwidth]{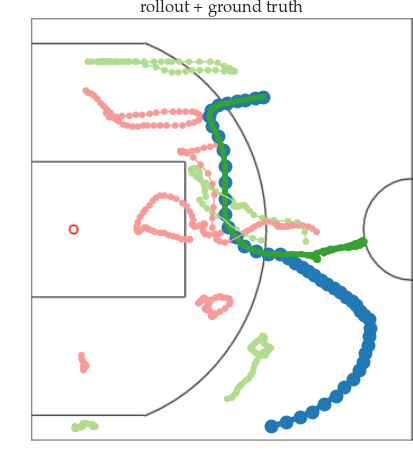}\\
\includegraphics[trim={20px 15px 0px 10px}, clip, width=\textwidth, height=1.1\textwidth]{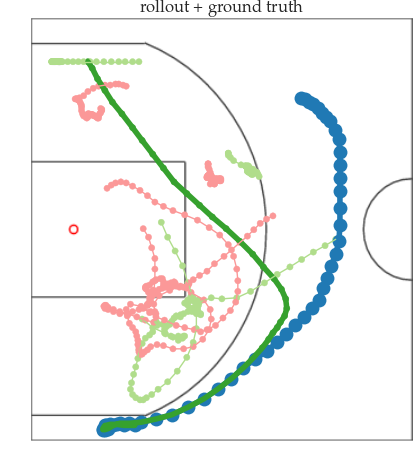}
\caption{\HPN rollouts\vspace{10pt}
}
\label{fig:macrogoals}
\end{subfigure}
\hspace{-1px}
\begin{subfigure}[b]{.19\textwidth}
 \centering
\includegraphics[trim={20px 15px 0px 10px}, clip, width=\textwidth, height=1.1\textwidth]{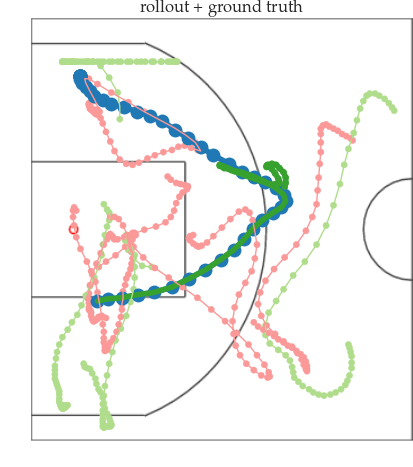}\\
\includegraphics[trim={20px 15px 0px 10px}, clip, width=\textwidth, height=1.1\textwidth]{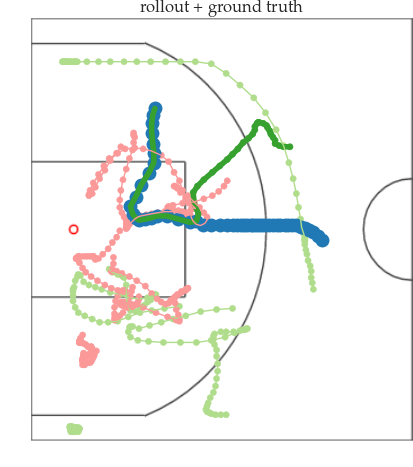}
\caption{\HPN rollouts\vspace{10pt}
}
\label{fig:macrogoals}
\end{subfigure}
\hspace{-1px}
\begin{subfigure}[b]{.19\textwidth}
 \centering
\includegraphics[trim={20px 15px 0px 10px}, clip, width=\textwidth, height=1.1\textwidth]{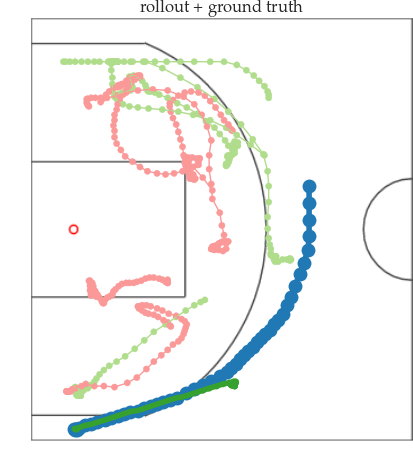}\\
\includegraphics[trim={20px 15px 0px 10px}, clip, width=\textwidth, height=1.1\textwidth]{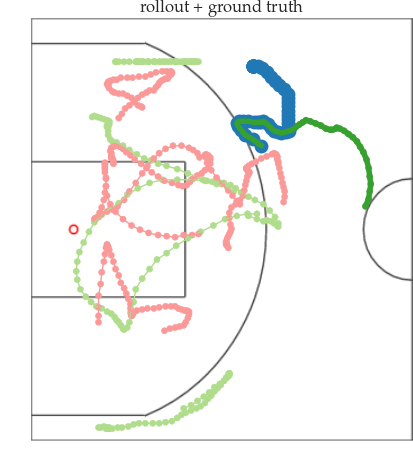}
\caption{\HPN rollouts\vspace{10pt}
}
\label{fig:macrogoals}
\end{subfigure}
\hspace{-1px}
\begin{subfigure}[b]{.19\textwidth}
 \centering
\includegraphics[trim={20px 15px 0px 10px}, clip, width=\textwidth, height=1.1\textwidth]{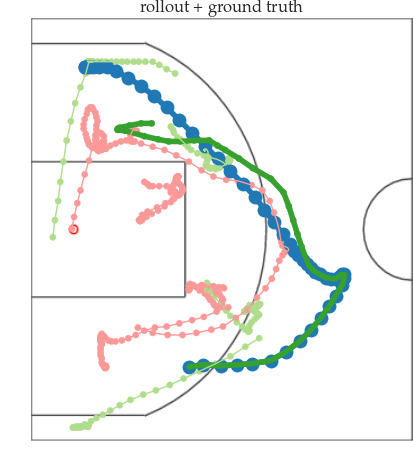}\\
\includegraphics[trim={20px 15px 0px 10px}, clip, width=\textwidth, height=1.1\textwidth]{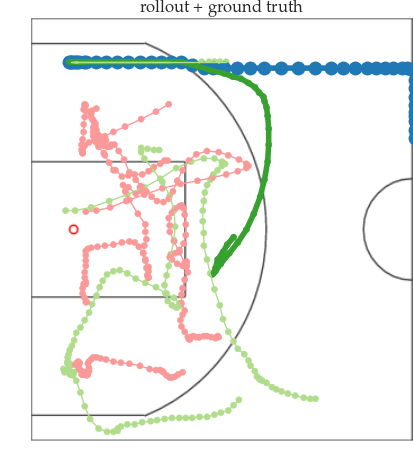}
\caption{\HPN (top) and failure case (bottom)\vspace{0pt}
}
\label{fig:macrogoals}
\end{subfigure}
\hspace{-1px}
\begin{subfigure}[b]{.19\textwidth}
 \centering
\includegraphics[trim={20px 15px 0px 10px}, clip, width=\textwidth, height=1.1\textwidth]{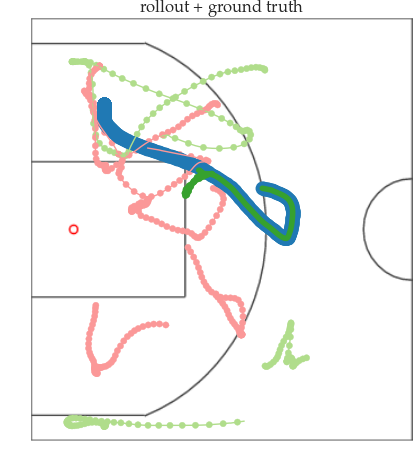}\\
\includegraphics[trim={20px 15px 0px 10px}, clip, width=\textwidth, height=1.1\textwidth]{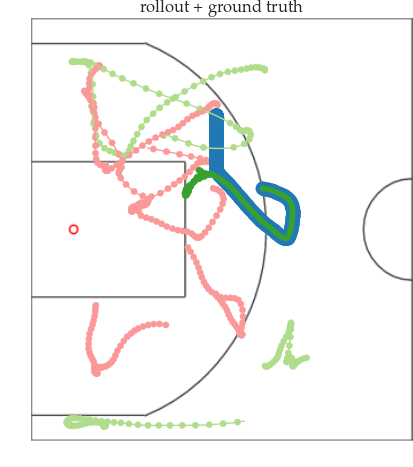}
\caption{\HPN (top), baseline (bottom)
}
\label{fig:macrogoals}
\end{subfigure}
\caption{Rollouts generated by the HPN (columns a, b, c, d) and baselines (column e). Each frame shows an offensive player (dark green), a rollout (blue) track that extrapolates after 20 frames, the offensive team (light green) and defenders (red). Note we do not show the ball, as we did not use semantic basketball features (i.e ``currently has the ball") during training.
The \HPN rollouts do not memorize training tracks (column a) and display a variety of natural behavior, such as curving, moving towards macro-goals and making sharp turns (c, bottom).
We also show a failure case (d, bottom), where the \HPN behaves unnaturally by moving along a straight line off the right side of the court -- which may be fixable by adding semantic game state information.
%
For comparison, a hierarchical baseline without an attention model, produces a straight-line rollout (column e, bottom), whereas the HPN produces a more natural movement curve (column e, top).
}
\label{fig:tracks}
\end{figure}

\subsection{How Realistic are the Generated Trajectories?}
\label{sec:qualitative-analysis}

The most holistic way to evaluate the trajectory rollouts is via visual analysis. Simply put, would a basketball expert find the rollouts by \HPN more realistic than those by the baselines? We begin by first visually analyzing some rollouts, and then present our human preference study results.

\textbf{Visualization.} Figure \ref{fig:tracks} depicts example rollouts for our HPN approach and one example rollout for a baseline.
Every rollout consists of two parts: 1) an initial ground truth phase from the holdout set and 2) a continuation by either the \HPN or ground truth.
We note a few salient properties. First, the \HPN does not memorize tracks, as the rollouts differ from the tracks it has trained on. Second, the \HPN generates rollouts with a high dynamic range, e.g. they have realistic curves, sudden changes of directions and move over long distances across the court towards macro-goals. For instance, \HPN tracks do not move towards macro-goals in unrealistic straight lines, but often take a curved route, indicating that the policy balances moving towards macro-goals with short-term responses to the current state (see also Figure \ref{fig:macrogoals}). In contrast, the baseline model often generates more constrained behavior, such as moving in straight lines or remaining stationary for long periods of time.

\begin{table}[t]
\centering
\footnotesize
\begin{tabular}{|l|| cc|| cc|| cc| }
\hline
\tb{Model comparison}
& \tb{Experts}
&
& \tb{Non-Experts}
&
& \tb{All}
&
\\
& \tb{W/T/L}
& \tb{Avg Gain}
& \tb{W/T/L}
& \tb{Avg Gain}
& \tb{W/T/L}
& \tb{Avg Gain}
\\
\hline
\tsc{vs-}\cnn
& 21 / 0 / 4
& 0.68
& 15 / 9 / 1
& 0.56
& 21 / 0 / 4
& 0.68
\\
\hline
\tsc{vs-}\grucnn
& 21 / 0 / 4
& 0.68
& 18 / 2 / 5
& 0.52
& 21 / 0 / 4
& 0.68
\\
\hline
\tsc{vs-}\grucnnhiercc
& 22 / 0 / 3
& 0.76
& 21 / 0 / 4
& 0.68
& 21 / 0 / 4
& 0.68
\\
\hline
\tsc{vs-ground truth}
& 11 / 0 / 14
& -0.12
& 10 / 4 / 11
& -0.04
& 11 / 0 / 14
& -0.12\\
\hline
\end{tabular}
\caption{
Preference study results. We asked basketball experts and knowledgeable non-experts to judge the relative quality of policy rollouts. We compare \HPN with ground truth and 3 baselines: a memory-less (\cnn) and memory-full (\grucnn) micro-planner and a hierarchical planner without attention (\grucnn\tsc{-cc}). For each of 25 test cases, \HPN wins if more judges preferred the \HPN rollout over a competitor. Average gain is the average signed vote (1 for always preferring \HPN, and -1 for never preferring). We see that the \HPN is preferred over all baselines (all results against baselines are 95\% statistically significant). Moreover, the \HPN is competitive with ground truth, indicating that \HPN generates realistic trajectories within our rollout setting. Please see the supplementary material for more details.
}
\label{tab:userstudy}
\end{table}

\textbf{Human preference study.} Our primary empirical result is a preference study eliciting judgments on the relative quality of rollout trajectories between \HPN and baselines or ground truth. We recruited seven experts (professional sports analysts) and eight knowledgeable non-experts (e.g., college basketball players) as judges.
Because all the learned policies perform better with a ``burn-in'' period, we first animated with the ground truth for 20 frames, and then extrapolated with a policy for 30 frames. During extrapolation, the other nine players do not animate.
For each test case, the users were shown an animation of two rollout extrapolations of a specific player's movement: one generated by the HPN, the other by a baseline or ground truth. The judges then chose which rollout looked more realistic. Please see the supplementary material for details of the study.

Table \ref{tab:userstudy} shows the preference study results. We tested 25 scenarios (some corresponding to scenarios in Figure \ref{fig:macrogoals}). \HPN won the vast majority of comparisons against the baselines using expert judges, with slightly weaker preference using non-expert judges. \HPN was also competitive with the ground truth. These results suggest that the \HPN is able to generate high-quality trajectories that significantly improve on baselines, and approach the ground truth quality for our extrapolation setting.
\begin{figure}[t!]
\centering
\begin{subfigure}[m]{.49\textwidth}
 \centering
\includegraphics[trim={39px 30px 0px 35px}, clip, width=\attwidth, height=\attheight]{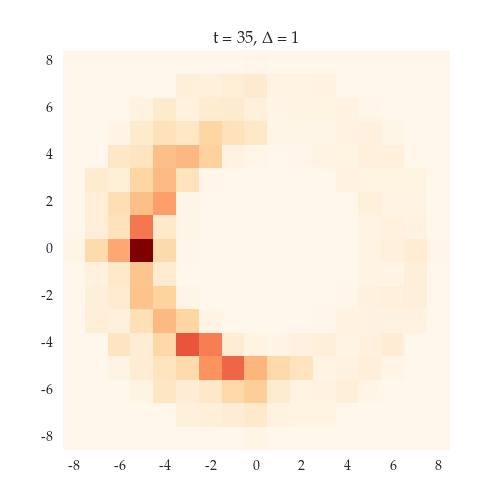}\hspace{-7pt}
\includegraphics[trim={39px 30px 0px 35px}, clip, width=\attwidth, height=\attheight]{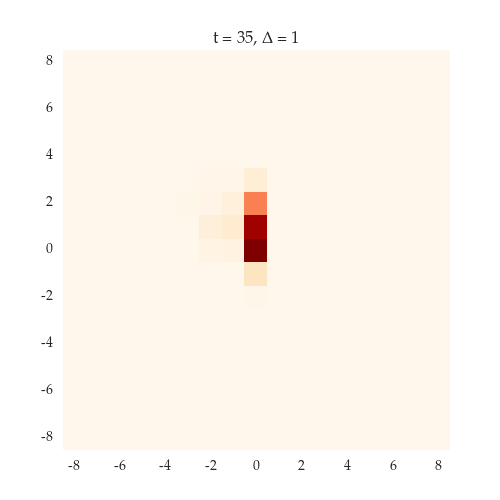}\hspace{-7pt}
\includegraphics[trim={39px 30px 0px 35px}, clip, width=\attwidth, height=\attheight]{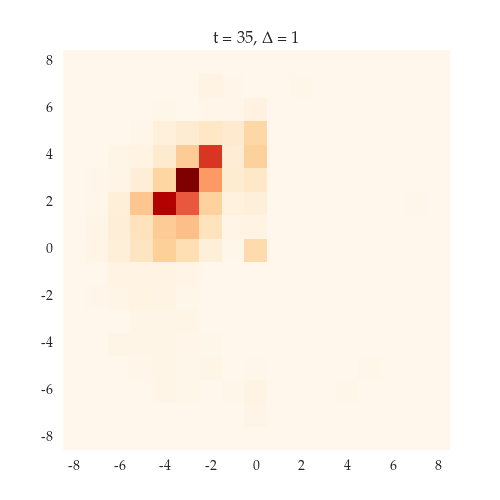}\\[-1pt]
\includegraphics[trim={39px 30px 0px 35px}, clip, width=\attwidth, height=\attheight]{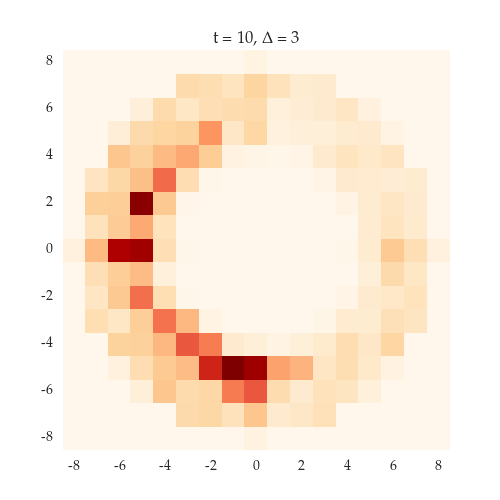}\hspace{-7pt}
\includegraphics[trim={39px 30px 0px 35px}, clip, width=\attwidth, height=\attheight]{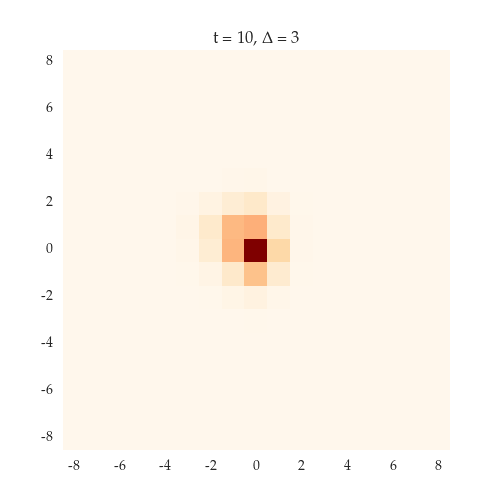}\hspace{-7pt}
\includegraphics[trim={39px 30px 0px 35px}, clip, width=\attwidth, height=\attheight]{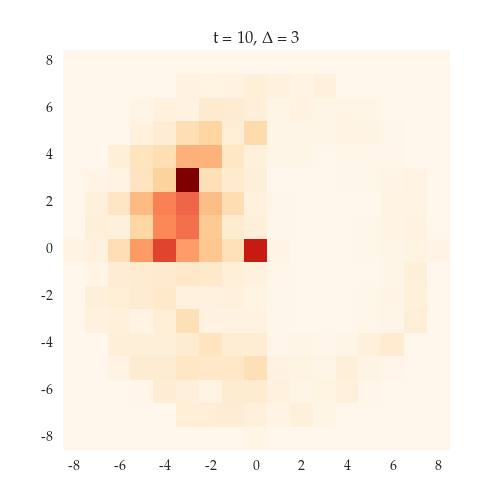}\\[-1pt]
\includegraphics[trim={39px 30px 0px 35px}, clip, width=\attwidth, height=\attheight]{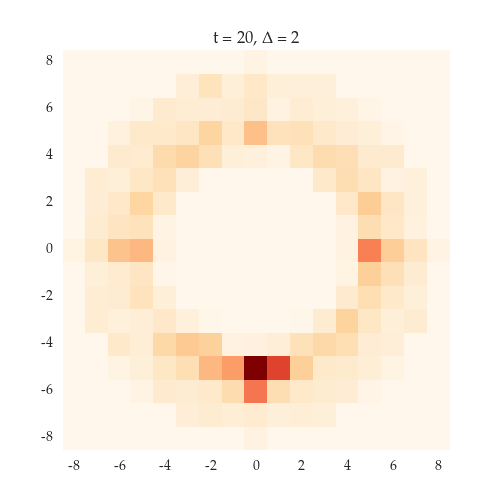}\hspace{-7pt}
\includegraphics[trim={39px 30px 0px 35px}, clip, width=\attwidth, height=\attheight]{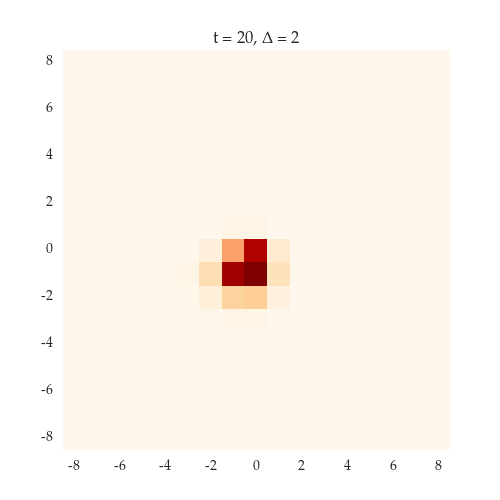}\hspace{-7pt}
\includegraphics[trim={39px 30px 0px 35px}, clip, width=\attwidth, height=\attheight]{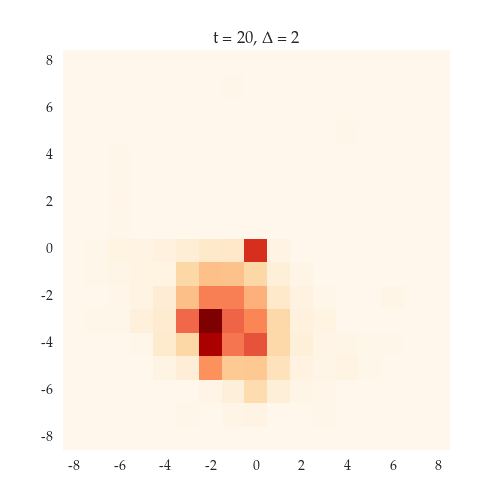}\\[-1pt]
\includegraphics[trim={39px 30px 0px 35px}, clip, width=\attwidth, height=\attheight]{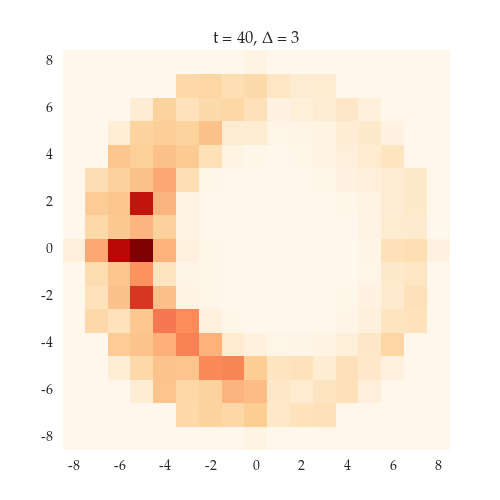}\hspace{-7pt}
\includegraphics[trim={39px 30px 0px 35px}, clip, width=\attwidth, height=\attheight]{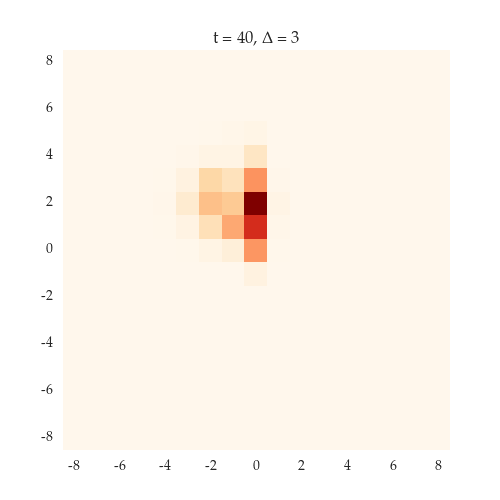}\hspace{-7pt}
\includegraphics[trim={39px 30px 0px 35px}, clip, width=\attwidth, height=\attheight]{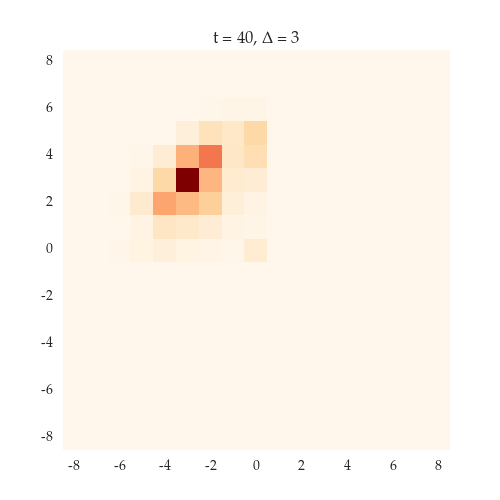}\\[-1pt]
\includegraphics[trim={39px 30px 0px 35px}, clip, width=\attwidth, height=\attheight]{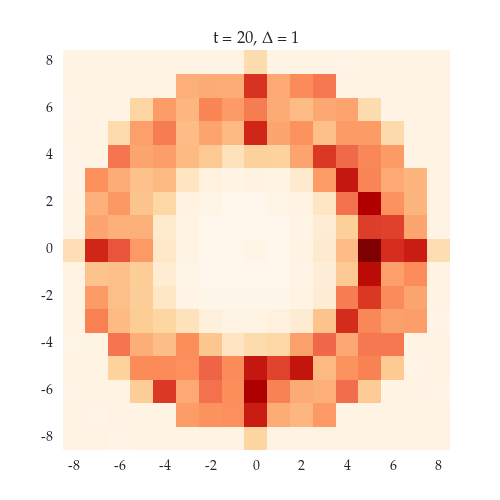}\hspace{-7pt}
\includegraphics[trim={39px 30px 0px 35px}, clip, width=\attwidth, height=\attheight]{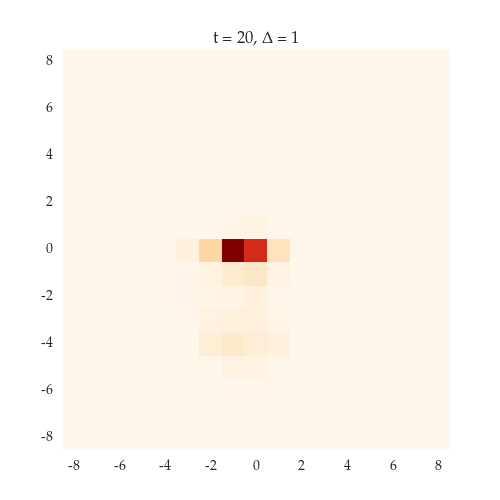}\hspace{-7pt}
\includegraphics[trim={39px 30px 0px 35px}, clip, width=\attwidth, height=\attheight]{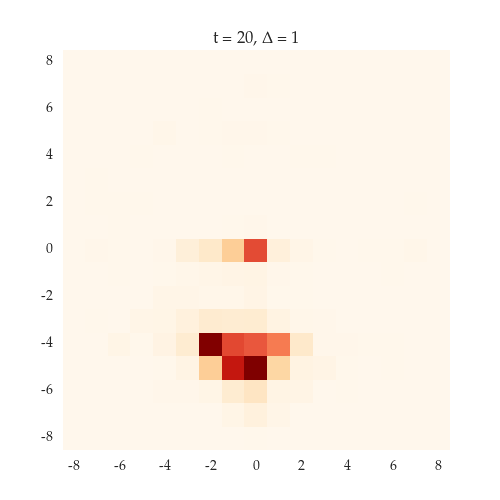}
\caption{
Predicted distributions for attention masks $\transferF$ (left column), velocity (micro-action) $\MicroPol$ (middle) and weighted velocity $\transferF(g) \odot \MicroPol$ (right) for basketball players. The center corresponds to 0 velocity.
}
\label{fig:attention}
\end{subfigure}%
\hspace{5pt}
\begin{subfigure}[m]{.49\textwidth}
 \centering
  \includegraphics[trim={20px 15px 0px 10px}, clip, width=0.47\textwidth, height=0.52\textwidth]{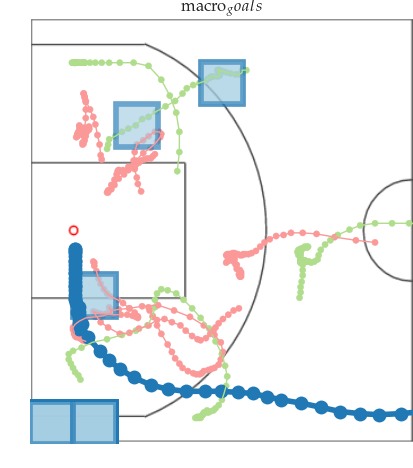}\hspace{-1pt}
  \includegraphics[trim={20px 15px 0px 10px}, clip, width=0.47\textwidth, height=0.52\textwidth]{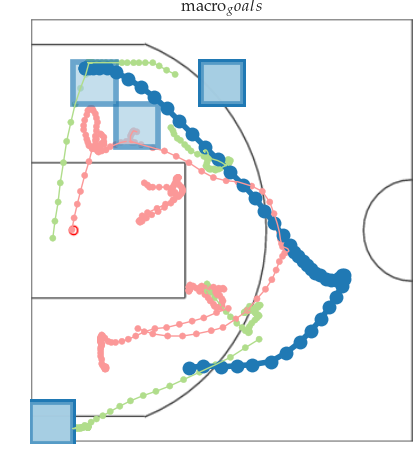} \\
  \includegraphics[trim={20px 15px 0px 10px}, clip, width=0.47\textwidth, height=0.52\textwidth]{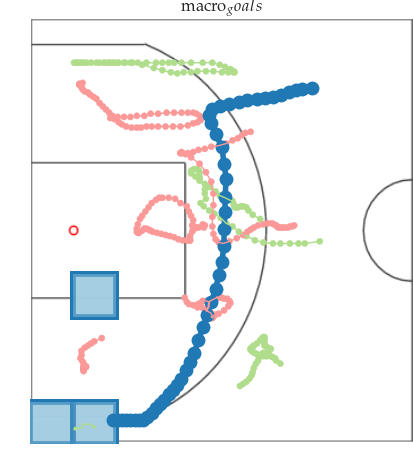}\hspace{-1pt}
  \includegraphics[trim={20px 15px 0px 10px}, clip, width=0.47\textwidth, height=0.52\textwidth]{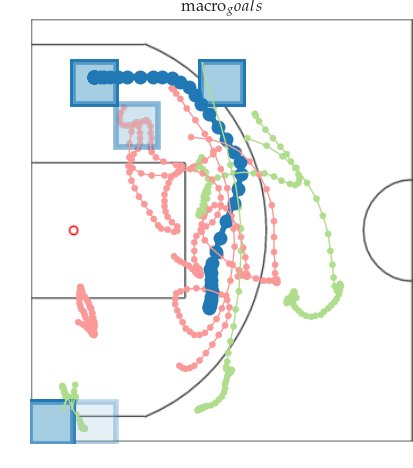}
\caption{Rollout tracks and predicted macro-goals $\GoalState_t$ (blue boxes). The \HPN starts the rollout after 20 frames. Macro-goal box intensity corresponds to relative prediction frequency during the trajectory.
}
\label{fig:macrogoals}
\end{subfigure}
 \caption{
 \textbf{Left:}
Visualizing how the attention mask generated from the macro-planner interacts with the micro-planner $\MicroPol$.
Row 1 and 2: the micro-planner $\MicroPol$ decides to stay stationary, but the attention $\transferF$ goes to the left. The weighted result $\transferF\odot\MicroPol$ is to move to the left, with a magnitude that is the average.
Row 3) $\MicroPol$ wants to go straight down, while $\transferF$ boosts the velocity so the agent bends to the bottom-left.
Row 4) $\MicroPol$ goes straight up, while $\transferF$ goes left: the agent goes to the top-left.
Row 5) $\MicroPol$ remains stationary, but $\transferF$ prefers to move in any direction. As a result, the agent moves down.
\textbf{Right:} The \HPN dynamically predicts macro-goals and guides the micro-planner in order to reach them. The macro-goal predictions are stable over a large number of timesteps. The \HPN sometimes predicts inconsistent macro-goals. For instance, in the bottom right frame, the agent moves to the top-left, but still predicts the macro-goal to be in the bottom-left sometimes.}
\label{fig:attention_and_macro_goals}
\vspace{-15pt}
\end{figure}

\subsection{Analyzing Macro- and Micro-planner Integration}
Our model integrates the macro- and micro-planner by converting the macro-goal into an attention mask on the micro-action output space, which intuitively guides the micro-planner towards the macro-goal. We now inspect our macro-planner and attention mechanism to verify this behavior.

Figure \ref{fig:attention} depicts how the macro-planner $\MacroPol$ guides the micro-planner $\MicroPol$ through the attention $\transferF$, by attending to the direction in which the agent can reach the predicted macro-goal. The attention model and micro-planner differ in semantic behavior: the attention favors a wider range of velocities and larger magnitudes, while the micro-planner favors smaller velocities. 

Figure \ref{fig:macrogoals} depicts predicted macro-goals by \HPN along with rollout tracks. In general, we see that the rollouts are guided towards the predicted macro-goals. However, we also observe that the \HPN makes some inconsistent macro-goal predictions, which suggests there is still ample room for improvement. \tempnewpage
\subsection{Benchmark Analysis}
\label{sec:quantitative-results}

We finally evaluated using a number of benchmark experiments, with results shown in  Table \ref{tab:shortterm}.  These experiments measure quantities that are related to overall quality, albeit not holistically.  We first evaluated the $4$-step look-ahead accuracy of the \HPN for micro-actions $a^i_{t+\Delta}, \Delta = 0,1,2,3$.
On this benchmark, the \HPN outperforms all baseline policy networks when not using weak labels $\hat{\transferF}$ to train the attention mechanism, which suggests that using a hierarchical model can noticeably improve the short-term prediction accuracy over non-hierarchical baselines.

We also report the prediction accuracy on weak labels $\hat{\GoalState},\hat{\transferF}$, although they were only used during pre-training, and we did not fine-tune for accuracy on them. Using weak labels one can tune the network for both long-term and short-term planning, whereas all non-hierarchical baselines are optimized for short-term planning only. Including the weak labels $\hat{\transferF}$ can lower the accuracy on short-term prediction, but increases the quality of the on-policy rollouts. This trade-off can be empirically set by tuning the number of weak labels used during pre-training.

\begin{table}[t]
\centering
\footnotesize
\begin{tabular}{|l | cccc | c|c |}
\hline
Model & $\Delta= 0$ & $\Delta= 1$ & $\Delta= 2$ & $\Delta= 3$ & Macro-goals $\GoalState$ & Attention $\transferF$  \\
\hline
\hline
\cnn  & 21.8\%& 21.5\%& 21.7\%& 21.5\% & -&-\\
\hline
\grucnn & 25.8\%& 25.0\%& 24.9\%& 24.4\% & -&-\\
\hline
\grucnnhiercc & 31.5\%& 29.9\%& 29.5\%& 29.1\% & 10.1\% &-\\
\hline
\grucnnhierstack & 26.9\%& 25.7\%& 25.9\%& 24.9\%& 9.8\% &-\\
\hline
\grucnnhierfull & \tb{33.7\%} & \tb{31.6\%} & \tb{31.0\%}& \tb{30.5\%} & 10.5\% &-\\
\hline
\grucnnhieraux & 31.6\%& 30.7\%& 29.4\%& 28.0\% & 10.8\% & 19.2\% \\
\hline
\end{tabular}
\caption{Benchmark Evaluations. $\Delta$-step look-ahead prediction accuracy for micro-actions $a^i_{t+\Delta}=\pi(\GameState_t)$ on a holdout set of player tracks, with $\Delta=0,1,2,3$. \grucnnhierstack is an \HPN where the final predictions are organized in a feed-forward stack, with $\pi(\GameState_t)_t$ feeding into $\pi(\GameState_t)_{t+1}$. Using attention (\grucnnhierfull) improves on all baselines in micro-action prediction.
All hierarchical models are pre-trained, but not fine-tuned, on macro-goals $\hat{\GoalState}$. We report empirical prediction accuracy on the weak labels $\hat{\GoalState}, \hat{\transferF}$ for hierarchical models.%
\grucnnhieraux is an \HPN that has also been trained on $\hat{\transferF}$. As $\hat{\transferF}$ optimizes for optimal long-term behavior, this lowers the micro-action prediction accuracy.
}
\label{tab:shortterm}
\end{table} \tempnewpage

\section{Limitations and Future Work}
We have presented a hierarchical policy class for generating long-term trajectories that models both macro-goals and micro-actions, and integrates them using a novel attention mechanism. We showed significant improvement over non-hierarchical baselines in a case study on basketball player behavior.

There are several notable limitations to our \HPN model. First, we did not consider all aspects of basketball gameplay, such as passing and shooting. We also modeled all players using a single policy whereas in reality player behaviors vary (although the variability can be low-dimensional \cite{Yue2014}). We also only modeled offensive players, and another interesting direction is modeling defensive players and integrating adversarial reinforcement learning \cite{panait2005cooperative} into our approach. These issues limited the scope of the rollouts in our preference study, and it would be interesting to consider extended settings.

In order to focus on our \HPN model class, we considered only the imitation learning setting. More broadly, many planning problems require collecting training data via exploration \cite{mnih_human-level_2015}, which can be more challenging. One interesting scenario is having two adversarial policies learn to be strategic against each other through repeatedly game-play in a basketball simulator. Furthermore, in general it can be difficult to acquire the appropriate weak labels to initialize the macro-planner training.

From a technical perspective, using attention in the output space may be applicable to other architectures. More sophisticated applications may require multiple levels of output attention masking.

\newpage

\bibliographystyle{unsrt}
\bibliography{litreview-1}


\end{document}